\documentclass{article}
\usepackage{spconf,amsmath,graphicx,amsfonts,bm}
\usepackage{amssymb}    
\usepackage{hyperref}
\usepackage{url}
\usepackage{multicol}
\usepackage{multirow}
\usepackage{booktabs}
\usepackage{algorithm}
\usepackage{algpseudocode}
\makeatletter
\def\BState{\State\hskip-\ALG@thistlm}
\makeatother

\title{Diversifying Deep Ensembles: A Saliency Map Approach for Enhanced OOD Detection, Calibration, and Accuracy}

\name{Stanislav Dereka\textsuperscript{$\dagger\ddagger$} \qquad Ivan Karpukhin\textsuperscript{$\dagger$} \qquad Maksim Zhdanov\textsuperscript{$\dagger\star$} \qquad Sergey Kolesnikov\textsuperscript{$\dagger$}}
\address{\textsuperscript{$\dagger$}Tinkoff \quad\textsuperscript{$\ddagger$}Moscow Institute of Physics and Technology \\ \textsuperscript{$\star$}National University of Science and Technology MISIS}

\begin{document}
\maketitle
\begin{abstract}
Deep ensembles are capable of achieving state-of-the-art results on classification and out-of-distribution (OOD) detection tasks. However, their effectiveness is limited due to the homogeneity of learned patterns within ensembles. To overcome this issue, our study introduces Saliency-Diversified Deep Ensembles (SDDE\footnote{Code implementation: \href{https://github.com/corl-team/sdde}{https://github.com/corl-team/sdde}.}), a novel approach that promotes diversity among ensemble members by leveraging saliency maps. Through incorporating saliency map diversification, our method outperforms conventional ensemble techniques and improves calibration on multiple classification and OOD detection tasks. In particular, the proposed method achieves state-of-the-art OOD detection quality, calibration, and accuracy on multiple benchmarks, including CIFAR10/100 and large-scale ImageNet datasets.
\end{abstract}


\section{Introduction}
In recent years, deep neural networks achieved state-of-the-art results on many computer vision tasks, including object detection \cite{liu2020detection}, classification \cite{he2015relu},
and face recognition \cite{deng2019arcface}. In image classification in particular, DNNs have demonstrated results more accurate than what humans are capable of on several popular benchmarks, such as ImageNet \cite{he2015relu}. However, these benchmarks often source both training and testing data from a similar distribution, while real-world scenarios frequently feature test sets curated independently and under varying conditions \cite{malinin2022shifts}. This disparity, known as domain shift, can have a significant negative impact on the performance of DNNs \cite{koh2020wilds}.
As such, ensuring robust confidence estimation and out-of-distribution (OOD) detection is paramount for achieving risk-controlled recognition \cite{yang2022openood}.

There has been a substantial amount of research focused on confidence estimation and OOD detection in deep learning. Some authors consider calibration refinements within softmax classifications \cite{guo2017calibration}, while others examine the nuances of Bayesian training \cite{goan2020bayesian}. In these studies, ensemble methods that use DNNs stand out by achieving superior outcomes in both confidence estimation and OOD detection \cite{lakshminarayanan2017deepensemble, yang2022openood}. The results of these methods can be further improved by diversifying model predictions and adopting novel training paradigms \cite{shui2018ncl,pang2019adp,rame2021dice}. However, current research on this subject primarily focuses on diversifying the model output without also diversifying the feature space.

\begin{figure}[t]
\centering
\vspace{0.2cm}
\includegraphics[width=\linewidth]{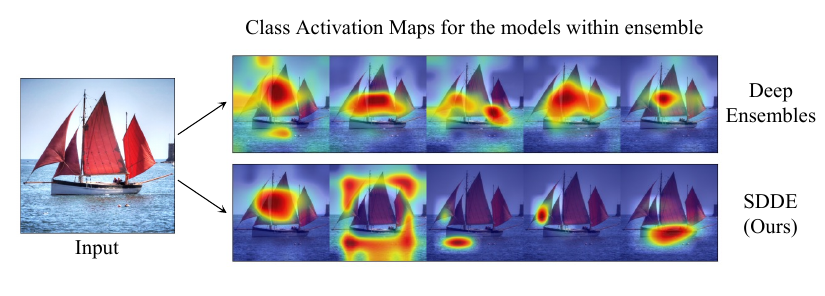}
\vspace{-0.8cm}
\caption{Saliency diversification. Compared to Deep Ensembles, the models within the proposed SDDE ensemble use different features for prediction, leading to improved generalization and confidence estimation.}
\label{fig:diversity}
\end{figure}

In this work, we introduce Saliency-Diversified Deep Ensembles (SDDE), a novel ensemble training method. Our approach encourages models to leverage distinct input features for making predictions, as shown in Figure \ref{fig:diversity}. This is implemented by computing saliency maps, or regions of interest, during the training process, and applying a special loss function for diversification. By incorporating these enhancements, we achieve new state-of-the-art (SOTA) results on multiple OpenOOD \cite{yang2022openood} benchmarks in terms of test set accuracy, confidence estimation, and OOD detection quality. Moreover, following past works \cite{hendrycks2018oe}, we extend our approach by adding OOD data to model training, which made it possible to obtain new SOTA results among methods that utilize OOD data during training.

The main contributions of our work are as follows:
\begin{enumerate}
    \item {We propose Saliency-Diversified Deep Ensembles (SDDE), a diversification technique that uses saliency maps in order to increase diversity among ensemble models.}
    \item{We achieve new SOTA results in OOD detection, calibration, and classification on the OpenOOD benchmark by using SDDE. In particular, we improve on OOD detection, accuracy, and calibration results on CIFAR10/100 datasets. On ImageNet-1K, we enhance the accuracy and OOD detection scores.}
    \item{We build upon our approach by adding OOD samples during ensemble training. The proposed method improves ensemble performance and establishes a new SOTA on the \mbox{CIFAR10} Near/Far and CIFAR100 Near OpenOOD benchmarks.}
\end{enumerate}

\section{Preliminaries}
In this section, we will give a brief description of the model and saliency maps estimation approaches used in the proposed SDDE method. Suppose $x \in \mathbb{R}^{CHW}$ is an input image with height $H$, width $W$, and $C$ channels. Let us consider a classification model $f(x) = h(g(x))$, which consists of a Convolutional Neural Network (CNN) $g: \mathbb{R}^{CHW} \rightarrow \mathbb{R}^{C'H'W'}$ and classifier $h: \mathbb{R}^{C'H'W'} \rightarrow \mathbb{R}^{L}$ with $L$ equal to the number of classes. CNN produces $C'$ feature maps $M^c, c=\overline{1, C'}$ with spatial dimensions $H'$ and $W'$. The output of the classifier is a vector of logits that can be mapped to class probabilities by a softmax layer.

The most common way to evaluate the importance of image pixels for output prediction is by computing saliency maps. In order to build a saliency map of the model, the simplest way is to compute output gradients w.r.t. input features \cite{simonyan2013saliencegrad}: 
\begin{equation}
    S_{Inp}(x) = \sum\limits_{i=1}^L\nabla_xf_i(x).
\label{eq:inp-grad}
\end{equation}

However, previous works show that input gradients can produce noisy outputs \cite{simonyan2013saliencegrad}. A more robust approach is to use Class Activation Maps (CAM) \cite{zhou2016cam}, or their generalization called GradCAM \cite{selvaraju2017gradcam}. Both methods utilize the spatial directions of the structure of CNNs with bottleneck features. Unlike CAM, GradCAM can be applied at any CNN layer, which is useful for small input images. In GradCAM, the region of interest is estimated by analyzing activations of the feature maps. The weight of each feature map channel is computed as
\begin{equation}
    \alpha_c = \frac{1}{H' W'}\sum\limits_{i = 1}^{H'}\sum\limits_{j = 1}^{W'} \frac{\delta f_y(x)}{\delta M^c_{i,j}},
\end{equation}
where $y \in \overline{1, L}$ is an image label. Having the weights of the feature maps, the saliency map is computed as
\begin{equation}
    S_{GradCAM}(x, y) = ReLU\left(\sum\limits_{c = 1}^{C'}\alpha_c M^c\right),
\end{equation}
where ReLU performs element-wise maximum between input and zero. The size of CAM is equal to the size of the feature maps. When visualized, CAMs are resized to match the size of the input image.

\section{Saliency Maps Diversity and Ensemble Agreement}\label{sec:toy}

\begin{figure}[h]
\centering
\vspace{-0.1in}
\includegraphics[width=0.9\linewidth]{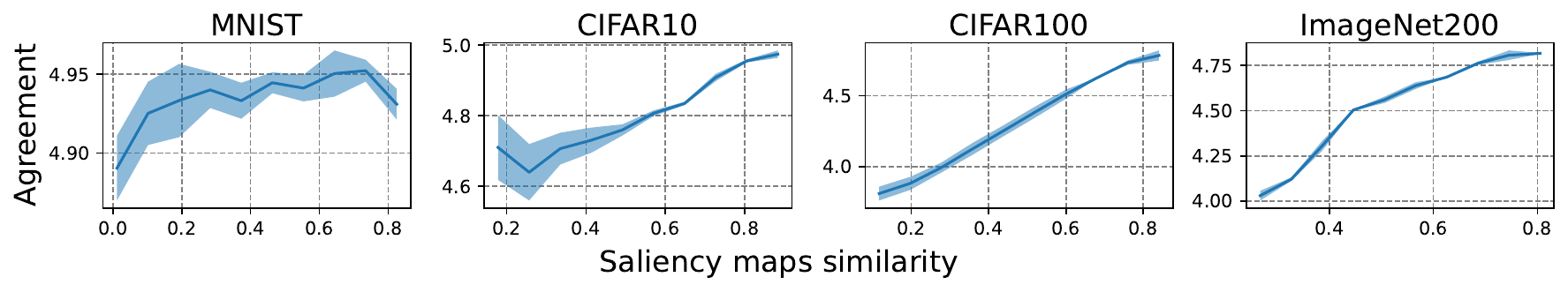}
\vspace{-0.15in}
\caption{The dependency of ensemble predictions agreement on saliency maps cosine similarity. Saliency maps are computed using GradCAM. Mean and STD values w.r.t. multiple training seeds are reported.}
\label{fig:motivation}
\end{figure}

Most previous works train ensemble models independently or by diversifying model outputs. Both of these approaches do not take into account the intrinsic algorithms implemented by the models. One way to distinguish classification algorithms is to consider the input features they use. Using saliency maps, which highlight input regions with the largest impact on the model prediction, is a popular technique for identifying these features. In this work, we suggest that ensemble diversity is related to the diversity of saliency maps. To validate this hypothesis, we analyzed the predictions and cosine similarity between the saliency maps of the Deep Ensemble model \cite{lakshminarayanan2017deepensemble} trained on the MNIST, CIFAR10, CIFAR100, and ImageNet200 datasets. In particular, we compute agreement as the number of models which are consistent with the ensemble prediction. As shown in Figure~\ref{fig:motivation}, higher saliency maps similarity usually leads to a larger agreement between models.

Given this observation, we raise the question of whether classification algorithms can be diversified by improving the diversity of saliency maps. We answer this question affirmatively by proposing a new SDDE method for training ensembles. In our experiments, we demonstrate the effectiveness of SDDE training in producing diverse high-quality models. 

\section{Method}\label{sec:method} 

\subsection{Saliency Maps Diversity Loss}
In Deep Ensembles \cite{lakshminarayanan2017deepensemble}, the models $f(x; \theta_k), k \in \overline{1, N}$ are trained independently. While there is a source of diversity in weight initialization, this method does not force different models to use different features. Therefore, we turn to the question of how we can train and implement different classification logic for models that rely on different input features. We answer this question by proposing a new loss function, which is motivated by previous research on saliency maps \cite{selvaraju2017gradcam}. 

The idea behind SDDE is to make the saliency maps \cite{simonyan2013saliencegrad} of the models as different as possible, as shown in Figure \ref{fig:diversity}. Suppose that we have computed a saliency map $S(x, y; \theta_k)$ for each model. The similarity of these models can be measured as a mean cosine similarity between their saliency maps. We thus propose the diversity loss function by computing the mean cosine similarity between the saliency maps of different models:
\begin{equation}
    \mathcal{L}_{div} (x, y; \theta) = \frac{2}{N(N-1)}\sum\limits_{k_1 > k_2}\left\langle S(x, y; \theta_{k_1}), S(x, y; \theta_{k_2}) \right\rangle.
\end{equation}
In SDDE, we use GradCAM \cite{selvaraju2017gradcam}, as it is more stable and requires less computation compared to the input gradient method \cite{simonyan2013saliencegrad}. As GradCAM is differentiable, the diversity loss and cross-entropy loss can be optimized together via gradient descent. The final loss has the following formula:
\begin{equation}
    \mathcal{L}(x, y; \theta) = \lambda \mathcal{L}_{div}(x, y; \theta) + \frac{1}{N} \sum\limits_k \mathcal{L}_{CE}(x, y; \theta_k),
\label{eq:sddeloss}
\end{equation}
where $\mathcal{L}_{CE}(x, y; \theta_k)$ is standard cross-entropy loss and $\lambda$ is a diversity loss weight.

\begin{figure}[t]
\centering
\includegraphics[width=\linewidth]{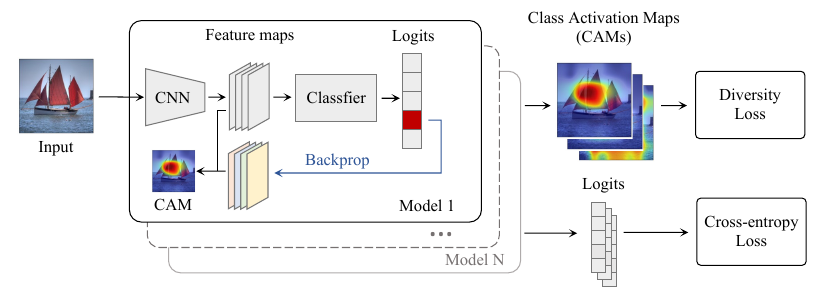}
\vspace{-0.3in}
\caption{The training pipeline, where we compute saliency maps using the GradCAM method for each model and apply a diversity loss. The final loss also includes the cross-entropy loss.}
\label{fig:loss}
\end{figure}

\subsection{Aggregation Methods for OOD Detection}
The original Deep Ensembles approach computes the average over softmax probabilities during inference. The naive approach to OOD detection is to use the Maximum Softmax Probability (MSP) \cite{hendrycks2016msp}. Suppose that there is an ensemble $f(x; \theta_k), k=\overline{1, N}$, where each model $f(x; \theta_k)$ maps input data to the vector of logits. Let us denote softmax outputs for each model as $p^k_i$ and average probabilities as $p_i$:
\begin{equation}
    p^k_i(x) = \frac{e^{f_i(x; \theta_k)}}{\sum\limits_j e^{f_j(x; \theta_k)}} = \mathrm{Softmax}_i(f(x; \theta_k)),
\end{equation}
\begin{equation}
    p_i(x)_ = \frac{1}{N} \sum\limits_{k}{p^k_i(x)}.
\end{equation}
The rule of thumb in ensembles is to predict the OOD score based on the maximum output probability:
\begin{equation}
    U_{MSP}(x) = \max\limits_i p_i(x).
\end{equation}

Certain recent works follow this approach \cite{abe2022necessary}.
However, other works propose to use logits instead of probabilities for OOD detection \cite{hendrycks2019mls}. In this work, we briefly examine which aggregation is better for ensembles by proposing the Maximum Average Logit (MAL) score. The MAL score extends the maximum logit approach for ensembles and is computed as:
\begin{equation}
    U_{MAL}(x) = \max\limits_i \frac{1}{N} \sum\limits_{k} f_i(x; \theta_k).
\end{equation}

\section{Experiments}

\subsection{Experimental Setup} \label{sec:details}

In our experiments, we follow the experimental setup and training procedure from the OpenOOD benchmark \cite{yang2022openood}. 
We use ResNet18 \cite{he2016resnet} for the CIFAR10, CIFAR100 \cite{krizhevsky2009cifar}, 
and ImageNet-200 datasets, LeNet \cite{lecun1998mnist} for the MNIST dataset \cite{lecun1998mnist}, and ResNet50 for the ImageNet-1K dataset. All models are trained using the SGD optimizer with a momentum of 0.9. The initial learning rate is set to 0.1 for ResNet18 and LeNet, 0.001 for ResNet50, and then reduced to $10^{-6}$ with the Cosine Annealing scheduler \cite{loshchilov2016cosineann}. Training lasts 50 epochs on MNIST, 100 epochs on CIFAR10 and CIFAR100, 
30 epochs on ImageNet-1K, and 90 epochs on ImageNet-200. In contrast to OpenOOD, we perform a multi-seed evaluation with 5 random seeds and report the mean and STD values for each experiment. In ImageNet experiments, we evaluate with 3 seeds. We compare the proposed SDDE method with other ensembling and diversification approaches, namely Deep Ensembles (DE) \cite{lakshminarayanan2017deepensemble}, Negative Correlation Learning (NCL) \cite{shui2018ncl}, Adaptive Diversity Promoting (ADP) \cite{pang2019adp}, and DICE diversification loss \cite{rame2021dice}. In the default setup, we train ensembles of 5 models. In SDDE, we set the parameter $\lambda$ from Equation \ref{eq:sddeloss} to 0.1 for MNIST and 0.005 for CIFAR10/100, 
ImageNet-200, and ImageNet-1K.

\subsection{Ensemble Diversity}

\begin{figure}[t!]
\centering
\includegraphics[width=\linewidth]{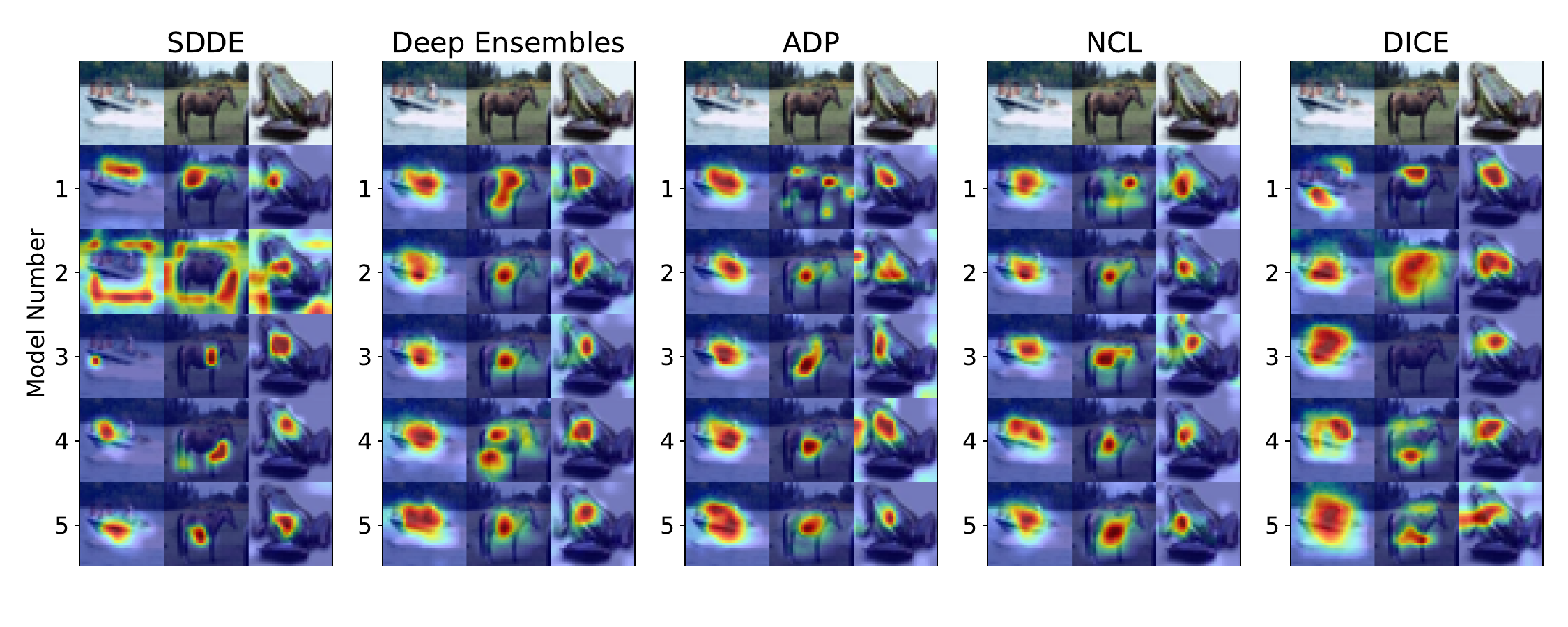}
\vspace{-0.3in}
\caption{Class Activation Maps (CAMs) for SDDE and the baseline methods. SDDE increases the diversity of CAMs by focusing on different regions of the images.}
\label{fig:cams}
\end{figure}
\begin{figure}[t]
\centering
\includegraphics[width=\linewidth]{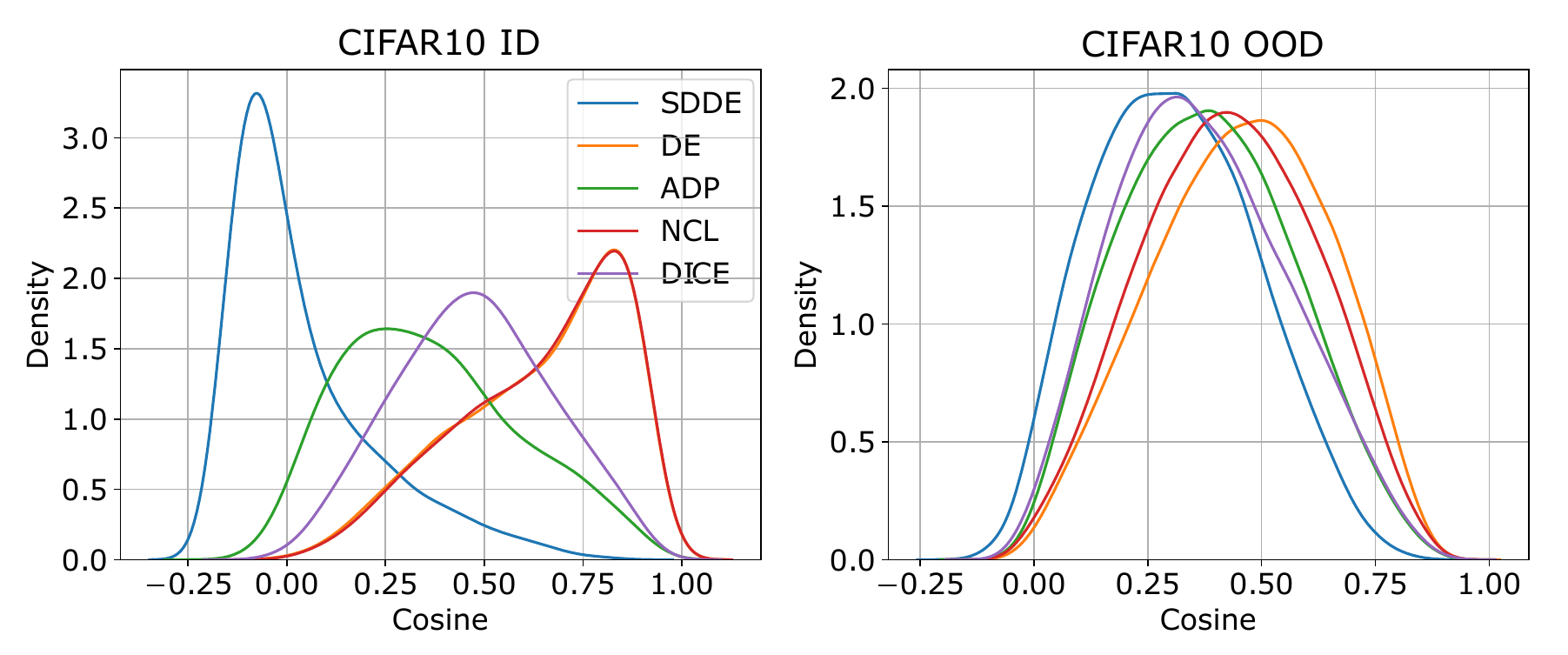}
\vspace{-0.3in}
\caption{Pairwise distributions of cosine similarities between Class Activation Maps (CAMs) of ensemble models.}
\label{fig:cos}
\end{figure}

To demonstrate the effectiveness of the proposed diversification loss, we evaluate the saliency maps for all considered methods, as shown in Figure \ref{fig:cams}.
The ground truth label is not available during inference, so we compute CAMs for the predicted class. It can be seen that SDDE models use different input regions for prediction. To highlight the difference between the methods, we analyze the pairwise cosine similarities between CAMs of different models in an ensemble. The distributions of cosine similarities are presented in Figure \ref{fig:cos}. According to the data, the baseline methods reduce similarity compared to Deep Ensemble. However, among all methods, SDDE achieves the lowest cosine similarity.
This effect persists even on OOD samples. In the following experiments, we study the benefits of the proposed diversification.

\textbf{Prediction-Based Diversity Metrics.}
Our saliency-based diversity approach primarily focuses on the variation of saliency maps, which may not fully capture the diversity in the prediction space. However, the benchmarked NCL, ADP, and DICE baselines mainly optimize the diversity in the prediction space. To fill this gap, we include additional diversity metrics, such as pairwise disagreement between networks, ratio-error, and Q-statistics from \cite{aksela2003comparison}. The values of metrics for SDDE and baseline methods are presented in Table \ref{tab:diversity}. It can be seen that SDDE has a lesser impact on the quality of individual models in the ensemble compared to other methods, while achieving moderate diversity values.

\begin{table}[h]
\centering
\caption{Diversification metrics. The best result in each column is \textbf{bolded}. The DICE method failed to converge on ImageNet.}
\resizebox{\linewidth}{!}{%
\begin{tabular}{@{}cc|ccc|cc@{}}
\toprule
\multirow{2}{*}{\bf Data} & \multirow{2}{*}{\bf Method} &\multicolumn{3}{c|}{\bf Diversity}& \multicolumn{2}{c}{\bf Error} \\
& & \bf Correlation$\downarrow$ & \bf Q-value$\downarrow$ & \bf D/S Rate $\uparrow$ & \bf Mean & \bf Ensemble \\
\hline
\multirow{5}{*}{\bf \parbox[t]{2mm}{{\rotatebox[origin=c]{90}{CIFAR 10}}}} 
 & DE & 58.85 \small $\pm$ 1.43 & 97.38 \small $\pm$ 0.31 & 0.22 \small $\pm$ 0.02 & 4.89 \small $\pm$ 0.13 & 3.93 \small $\pm$ 0.10 \\
 & NCL & 59.92 \small $\pm$ 0.81 & 97.53 \small $\pm$ 0.17 & 0.22 \small $\pm$ 0.01 & 4.95 \small $\pm$ 0.13 & 4.04 \small $\pm$ 0.08 \\
 & ADP & \bf 57.52 \small $\pm$ 0.39 & \bf 97.03 \small $\pm$ 0.09 & \bf 0.32 \small $\pm$ 0.01 & 5.14 \small $\pm$ 0.04 & 3.93 \small $\pm$ 0.12 \\
 & DICE & 57.61 \small $\pm$ 4.15 & 96.82 \small $\pm$ 1.32 & 0.25 \small $\pm$ 0.05 & 5.24 \small $\pm$ 0.57 & 4.10 \small $\pm$ 0.16 \\
 & SDDE & 59.29 \small $\pm$ 0.95 & 97.47 \small $\pm$ 0.18 & 0.23 \small $\pm$ 0.01 & \bf 4.87 \small $\pm$ 0.11 & \bf 3.92 \small $\pm$ 0.11 \\
\hline
\multirow{5}{*}{\bf \parbox[t]{2mm}{{\rotatebox[origin=c]{90}{CIFAR 100}}}}
 & DE & 62.90 \small $\pm$ 2.25 & 92.70 \small $\pm$ 1.42 & 0.86 \small $\pm$ 0.10 & 23.12 \small $\pm$ 1.28 & 19.05 \small $\pm$ 0.54 \\
 & NCL & 63.06 \small $\pm$ 1.32 & 92.86 \small $\pm$ 0.70 & 0.83 \small $\pm$ 0.04 & 23.00 \small $\pm$ 0.47 & 19.08 \small $\pm$ 0.20 \\
 & ADP & 63.02 \small $\pm$ 0.78 & 92.44 \small $\pm$ 0.40 & \bf 1.78 \small $\pm$ 0.08 & 25.13 \small $\pm$ 0.12 & 18.82 \small $\pm$ 0.06 \\
 & DICE & \bf 60.88 \small $\pm$ 0.41 & \bf 91.78 \small $\pm$ 0.26 & 0.98 \small $\pm$ 0.01 & 23.20 \small $\pm$ 0.25 & 18.74 \small $\pm$ 0.25 \\
 & SDDE & 61.43 \small $\pm$ 1.94 & 92.09 \small $\pm$ 1.13 & 0.86 \small $\pm$ 0.07 & \bf 22.80 \small $\pm$ 0.63 & \bf 18.67 \small $\pm$ 0.25 \\
 \hline
 \multirow{5}{*}{\bf \parbox[t]{2mm}{{\rotatebox[origin=c]{90}{\hspace{0.13in}ImageNet}}}}
 & DE & 93.65 \small $\pm$ 0.03 & 99.84 \small $\pm$ 0.00 & 0.09 \small $\pm$ 0.00 & 25.41 \small $\pm$ 0.00 & 24.84 \small $\pm$ 0.06\\
 & NCL & 99.31 \small $\pm$ 0.03 & 100.00 \small $\pm$ 0.00 & 0.01 \small $\pm$ 0.00 & 25.41 \small $\pm$ 0.11 & 24.95 \small $\pm$ 0.13\\
 & ADP & \bf 80.93 \small $\pm$ 0.67 & \bf 98.05 \small $\pm$ 0.10 & \bf 2.43 \small $\pm$ 0.04 & 34.56 \small $\pm$ 0.20 &  24.90 \small $\pm$ 0.01 \\
 & SDDE& 93.80 \small $\pm$ 0.09 & 99.85 \small $\pm$ 0.00 & 0.09 \small $\pm$ 0.00 & \bf 25.39 \small $\pm$ 0.00 & \bf 24.80 \small $\pm$ 0.01\\
\bottomrule
\end{tabular}%
}
\label{tab:diversity}
\end{table}

\subsection{Ensemble Accuracy and Calibration}

Ensemble methods are most commonly used for improving classification accuracy and prediction calibration metrics. We compare these aspects of SDDE with other ensemble methods by measuring the test set classification accuracy, Negative Log-Likelihood (NLL), Expected Calibration Error (ECE) \cite{guo2017calibration}
, and Brier score \cite{brier1950verification}. All metrics are computed after temperature tuning on the validation set \cite{guo2017calibration}. The results are presented in \mbox{Table \ref{tab:ensembles}}. It can be seen that the SDDE approach outperforms other methods on CIFAR10 and CIFAR100 in terms of both accuracy and calibration. 

\begin{table}[h]
\caption{Accuracy and calibration metrics. The best results for each dataset and metric are \textbf{bolded}.}
\centering
\resizebox{\linewidth}{!}{
\begin{tabular}{c|l|ccccc}
\toprule
\bf Data & \bf Metric & \bf DE & \bf NCL & \bf ADP & \bf DICE & \bf SDDE\\
\hline
\multirow{4}{*}{\bf \parbox[t]{2mm}{{\rotatebox[origin=c]{90}{MNIST}}}}& NLL \textsubscript{($\times10$)} $\downarrow$& 0.38 \small $\pm$ 0.02 & 0.36 \small $\pm$ 0.02 & \bf 0.35 \small $\pm$ 0.02 & 0.37 \small $\pm$ 0.02 & 0.37 \small $\pm$ 0.03 \\
& ECE \textsubscript{($\times10^2$)} $\downarrow$& 0.21 \small $\pm$ 0.07 & \bf 0.19 \small $\pm$ 0.07 & 0.31 \small $\pm$ 0.05 & 0.29 \small $\pm$ 0.07 & 0.28 \small $\pm$ 0.08 \\
& Brier \textsubscript{($\times 10^2$)} $\downarrow$& 1.95 \small $\pm$ 0.09 & 1.89 \small $\pm$ 0.08 & \bf 1.77 \small $\pm$ 0.12 & 1.84 \small $\pm$ 0.09 & 1.92 \small $\pm$ 0.15 \\
& Acc \textsubscript{(\%)} $\uparrow$& 98.72 \small $\pm$ 0.07 & 98.76 \small $\pm$ 0.03 & \bf 98.85 \small $\pm$ 0.11 & 98.81 \small $\pm$ 0.09 & 98.73 \small $\pm$ 0.14 \\
\hline
\multirow{4}{*}{\bf \parbox[t]{2mm}{{\rotatebox[origin=c]{90}{CIFAR10}}}}& NLL \textsubscript{($\times10$)} $\downarrow$& 1.30 \small $\pm$ 0.02 & 1.33 \small $\pm$ 0.01 & 1.36 \small $\pm$ 0.02 & 1.45 \small $\pm$ 0.03 & \bf 1.28 \small $\pm$ 0.01 \\
& ECE \textsubscript{($\times10^2$)} $\downarrow$& 0.84 \small $\pm$ 0.16 & 0.87 \small $\pm$ 0.13 & 1.00 \small $\pm$ 0.14 & 1.33 \small $\pm$ 0.06 & \bf 0.82 \small $\pm$ 0.15 \\
& Brier \textsubscript{($\times 10^2$)} $\downarrow$& 5.96 \small $\pm$ 0.08 & 6.03 \small $\pm$ 0.08 & 5.99 \small $\pm$ 0.07 & 6.33 \small $\pm$ 0.23 & \bf 5.88 \small $\pm$ 0.07 \\
& Acc \textsubscript{(\%)} $\uparrow$& 96.07 \small $\pm$ 0.10 & 95.96 \small $\pm$ 0.08 & 96.07 \small $\pm$ 0.12 & 95.90 \small $\pm$ 0.16 & \bf 96.08 \small $\pm$ 0.11 \\
\hline
\multirow{4}{*}{\bf \parbox[t]{2mm}{{\rotatebox[origin=c]{90}{CIFAR100}}}}& NLL \textsubscript{($\times10$)} $\downarrow$& 7.21 \small $\pm$ 0.14 & 7.22 \small $\pm$ 0.05 & 7.54 \small $\pm$ 0.03 & 7.46 \small $\pm$ 0.054 & \bf 6.93 \small $\pm$ 0.05 \\
& ECE \textsubscript{($\times10^2$)} $\downarrow$& 3.78 \small $\pm$ 0.17 & 3.86 \small $\pm$ 0.23 & 3.73 \small $\pm$ 0.28 & 4.46 \small $\pm$ 0.36 & \bf 3.48 \small $\pm$ 0.36 \\
& Brier \textsubscript{($\times 10^2$)} $\downarrow$& 27.33 \small $\pm$ 0.76 & 27.34 \small $\pm$ 0.33 & 27.08 \small $\pm$ 0.17 & 27.12 \small $\pm$ 0.26 & \bf 26.64 \small $\pm$ 0.34 \\
& Acc \textsubscript{(\%)} $\uparrow$& 80.95 \small $\pm$ 0.54 & 80.92 \small $\pm$ 0.20 & 81.18 \small $\pm$ 0.06 & 81.26 \small $\pm$ 0.25 & \bf 81.33 \small $\pm$ 0.28 \\
\bottomrule
\end{tabular}
}
\label{tab:ensembles}
\end{table}

\subsection{OOD Detection}\label{exp:ood}
We evaluate SDDE's OOD detection performance on the OpenOOD benchmark. Near-OOD datasets exhibit only a semantic shift compared to ID datasets, whereas far-OOD datasets show a significant covariate shift. Since SDDE does not use external data for training, we compare it to ensembles that only use in-distribution data. The results are presented in Table \ref{tab:ood}. It can be seen that SDDE achieves SOTA results in almost all cases, including the total scores on near and far tests.

\begin{table}[h]
\centering
\caption{OOD detection results. All methods are trained on the ID dataset and tested on multiple OOD sources. Mean and STD AUROC values are reported. The best results among different methods are \textbf{bolded}.}
\resizebox{0.7\linewidth}{!}{%
\begin{tabular}{@{}cc|cc@{}}
\toprule
\multicolumn{1}{c|}{\bf ID Dataset} & \bf Method& \multicolumn{1}{l|}{\bf OOD \textsubscript{Near}} & \bf OOD \textsubscript{Far}\\ \hline
\multicolumn{1}{c|}{\multirow{5}{*}{\bf MNIST}} & DE & \multicolumn{1}{l|}{92.45 \small $\pm$ 0.60} & 98.34 \small $\pm$ 0.31 \\
\multicolumn{1}{l|}{} & NCL & \multicolumn{1}{l|}{92.23 \small $\pm$ 0.53} & 98.56 \small $\pm$ 0.19 \\
\multicolumn{1}{l|}{} & ADP & \multicolumn{1}{l|}{93.74 \small $\pm$ 0.48} & 98.96 \small $\pm$ 0.32 \\
\multicolumn{1}{l|}{} & DICE & \multicolumn{1}{l|}{92.87 \small $\pm$ 0.67} & 98.76 \small $\pm$ 0.41 \\
\multicolumn{1}{l|}{} & \textbf{SDDE (Our)} & \multicolumn{1}{l|}{\bf 96.56 \small $\pm$ 0.67} & \bf 99.88 \small $\pm$ 0.02 \\ \hline
\multicolumn{1}{c|}{\multirow{5}{*}{\bf CIFAR10}} & DE & \multicolumn{1}{l|}{91.07 \small $\pm$ 0.17} & 93.55 \small $\pm$ 0.20 \\
\multicolumn{1}{l|}{} & NCL & \multicolumn{1}{l|}{91.22 \small $\pm$ 0.09} & 93.52 \small $\pm$ 0.17 \\
\multicolumn{1}{l|}{} & ADP & \multicolumn{1}{l|}{90.79 \small $\pm$ 0.17} & 93.50 \small $\pm$ 0.16 \\
\multicolumn{1}{l|}{} & DICE & \multicolumn{1}{l|}{89.99 \small $\pm$ 0.63} & 92.89 \small $\pm$ 0.47 \\
\multicolumn{1}{l|}{} & \textbf{SDDE (Our)} & \multicolumn{1}{l|}{\bf 92.06 \small $\pm$ 0.13} & \bf 94.60 \small $\pm$ 0.17 \\ \hline
\multicolumn{1}{c|}{\multirow{5}{*}{\bf CIFAR100}} & DE & \multicolumn{1}{l|}{82.55 \small $\pm$ 0.51} & 80.61 \small $\pm$ 0.51 \\
\multicolumn{1}{l|}{} & NCL & \multicolumn{1}{l|}{82.79 \small $\pm$ 0.18} & 80.53 \small $\pm$ 0.66 \\
\multicolumn{1}{l|}{} & ADP & \multicolumn{1}{l|}{82.98 \small $\pm$ 0.18} & 81.37 \small $\pm$ 0.54 \\
\multicolumn{1}{l|}{} & DICE & \multicolumn{1}{l|}{83.18 \small $\pm$ 0.14} & 82.15 \small $\pm$ 0.85 \\
\multicolumn{1}{l|}{} & \textbf{SDDE (Our)} & \multicolumn{1}{l|}{\bf 83.65 \small $\pm$ 0.09} & \bf 82.39 \small $\pm$ 0.56 \\
\bottomrule
\end{tabular}
}
\label{tab:ood}
\end{table}

\subsection{ImageNet Results}
We conduct experiments on the large-scale ImageNet-1K benchmark from OpenOOD in addition to the above-mentioned datasets. The accuracy, calibration, and OOD detection results are presented in Table \ref{tab:largescale}. It can be seen that SDDE achieves the best accuracy and OOD detection score among all the methods on ImageNet-1K. Furthermore, SDDE achieves calibration scores comparable to the best-performing method in each column. The DICE training failed to converge in this setup, and we excluded it from the comparison.

\begin{table}[h]
\centering
\caption{ImageNet results. The best-performing method in each column is \textbf{bolded}. 
}
\resizebox{\linewidth}{!}{%
\begin{tabular}{c|cccccc}
\toprule
 \textbf{Method} & \textbf{OOD\textsubscript{Near}}& \textbf{OOD\textsubscript{Far}}& \bf NLL \textsubscript{($\times10$)}& \bf ECE \textsubscript{($\times10^2$)}& \bf Brier \textsubscript{($\times10^2$)}& \textbf{Acc \textsubscript{(\%)}}\\
 \hline
 DE & 75.02 \small ± 0.06& 82.72 \small ± 0.26& 9.79 \small ± 0.00& 1.54 \small ± 0.06& \bf 34.40 \small ± 0.01& 75.16 \small ± 0.06\\
  NCL & 75.06 \small ± 0.05& 82.85 \small ± 0.17& 9.83 \small ± 0.00& \bf 1.51 \small ± 0.02& 34.51 \small ± 0.14& 75.05 \small ± 0.13\\
  ADP & \bf 75.26 \small ± 0.01& 82.62 \small ± 0.08 & 9.95 \small ± 0.01& 1.95 \small ± 0.11& 34.54 \small ± 0.08& 75.10 \small ± 0.01\\
  \bf SDDE & \bf 75.26 \small ± 0.02& \bf 85.98 \small ± 0.61& \bf 9.79 \small ± 0.01& 1.55 \small ± 0.05& 34.43 \small ± 0.05& \bf 75.20 \small ± 0.01\\
 \bottomrule
\end{tabular}%
}
\label{tab:largescale}
\end{table}

\subsection{Distribution Shifts}

In addition to OOD detection, accuracy, and calibration, we evaluate SDDE's performance on datasets with distribution shifts (OOD generalization), such as the CIFAR10-C and CIFAR100-C datasets \cite{hendrycks2019robustness}. The accuracy and calibration metrics are reported in Table~\ref{tab:corrupted}. These results demonstrate that SDDE outperforms the baselines in both accuracy and calibration metrics on CIFAR10-C, and achieves accuracy on par with the best-performing method while improving the calibration metrics on CIFAR100-C.

\begin{table}[h]
\caption{Accuracy and calibration metrics on corrupted datasets.}
\label{tab:corrupted}
\centering
\resizebox{0.8\linewidth}{!}{%
\begin{tabular}{c|cccc}
\toprule
\multirow{2}{*}{\bf Method} & \multicolumn{4}{c}{\bf CIFAR10-C}  \\
 & \bf NLL \textsubscript{($\times10$)}& \bf ECE \textsubscript{($\times10^2$)}& \bf Brier \textsubscript{($\times10^2$)}& \bf Accuracy \\
 \hline
DE& 1.06 \small ± 0.01& \bf 4.67 \small ± 0.12 & 44.68 \small ± 0.58& 67.74 \small ± 0.38 \\
NCL& 1.06 \small ± 0.01& 4.83 \small ± 0.37 & 44.98 \small ± 0.49& 67.43 \small ± 0.40 \\
ADP& 1.06 \small ± 0.02 & 4.67 \small ± 0.12 & 44.64 \small ± 0.58& 67.81 \small ± 0.38 \\
DICE& 1.10 \small ± 0.02 & 7.41 \small ± 0.53 & 45.86 \small ± 0.70& 67.14 \small ± 0.60 \\
\textbf{SDDE}& \bf 1.04 \small ± 0.01& 4.84 \small ± 0.38 & \bf 44.32 \small ± 0.31& \bf 67.96 \small ± 0.28 \\ \hline
 & \multicolumn{4}{c}{\bf CIFAR100-C}\\ \hline
 DE& 12.91 \small ± 0.10 & 4.74 \small ± 0.26 & 43.28 \small ± 0.59 & 68.55 \small ± 0.55 \\
 NCL& 12.83 \small ± 0.12 & 4.69 \small ± 0.10 & 43.07 \small ± 0.36 & 68.71 \small ± 0.30 \\
 ADP& 13.14 \small ± 0.11 & 4.97 \small ± 0.10 & 42.61 \small ± 0.32 & \textbf{69.20 \small ± 0.22} \\
 DICE& 13.57 \small ± 0.09 & 5.23 \small ± 0.26 & 44.40 \small ± 0.50 & 67.91 \small ± 0.48 \\
 \textbf{SDDE}& \textbf{12.58 \small ± 0.08} & \textbf{4.00 \small ± 0.09}& \textbf{42.43 \small ± 0.32} & 69.08 \small ± 0.27 \\ \bottomrule
\end{tabular}%
}
\end{table}

\subsection{Leveraging OOD Data for Training}
\label{exp:sdde_ood}
In some applications, an unlabeled OOD sample can be provided during training to further improve OOD detection quality.
According to previous studies \cite{yang2022openood}, Outlier Exposure (OE) \cite{hendrycks2018oe} is one of the most accurate methods on CIFAR10/100 and ImageNet-200 datasets. The idea behind OE is to make predictions on OOD data as close to the uniform distribution as possible. This is achieved by minimizing cross-entropy between the uniform distribution and model output on OOD data:
\begin{equation}
    \mathcal{L}_{OE}(x; \theta) = -\frac{1}{C}\sum\limits_i \log \mathrm{Softmax}_i(f(x; \theta)).
\end{equation}

Given an unlabeled OOD sample $\overline{x}$, we combine OE loss with SDDE, leading to the following objective: 
\begin{equation}
\mathcal{L}_{OOD}(x, y, \overline{x}; \theta) = \mathcal{L}(x, y; \theta) + \beta  \frac{1}{N} \mathcal{L}_{OE}(\overline{x}; \theta_k),
\end{equation}
We follow the original OE implementation and set $\beta$ to $0.5$. We call the final method SDDE\textsubscript{OOD}.

The OpenOOD implementation of OE includes only a single model. To have a fair comparison, we train an ensemble of 5 OE models and average their predictions.

The results are presented in Table \ref{tab:oe-new}. It can be seen that SDDE\textsubscript{OOD} achieves higher OOD detection accuracy compared to the current SOTA OE method in OpenOOD. Following SDDE experiments, we evaluate SDDE\textsubscript{OOD} calibration and accuracy. SDDE\textsubscript{OOD} demonstrates competitive accuracy and calibration scores on all benchmark datasets.

\begin{table}[t!]
\centering
\caption{Evaluation results for methods trained with OOD data.}
\resizebox{\linewidth}{!}{%
\begin{tabular}{ll|llllll@{}}
\toprule
   \multicolumn{2}{l|}{\multirow{2}{*}{\textbf{Method}}}& \multicolumn{6}{c}{\bf CIFAR-10}\\
 \multicolumn{2}{l|}{} & \textbf{OOD\textsubscript{Near}}& \textbf{OOD\textsubscript{Far}}& \textbf{NLL \small ($\times10$)}& \textbf{ECE \small ($\times10^2$)}& \textbf{Brier \small ($\times10^2$)}& \textbf{Accuracy \small (\%)}\\ \hline
 \multicolumn{2}{l|}{OE\textsubscript{Single}} & 94.82 \small 
± 0.21& 96.00 \small  ± 0.13& 2.85 \small  ± 0.26& 5.88 \small  ± 0.97& 11.62  \small ± 0.98& 94.63  \small ± 0.26\\
 \multicolumn{2}{l|}{OE\textsubscript{Ensemble}} & \textbf{96.23 \small  ± 0.08}& 97.45  \small ± 0.15& 2.65  \small ± 0.05& \textbf{4.86  \small ± 0.25}& 10.89  \small ± 0.19& 95.84  \small ± 0.19\\
 \multicolumn{2}{l|}{\textbf{SDDE\textsubscript{OOD}}} & 96.22 \small ± 0.08& \textbf{97.56  \small ± 0.07}& \textbf{2.64  \small ± 0.02}& 4.95 \small  ± 0.15& \textbf{10.86  \small ± 0.15}& \textbf{95.87  \small ± 0.02}\\ \hline
 & & \multicolumn{6}{c}{\bf CIFAR-100}\\ \hline
 \multicolumn{2}{l|}{OE\textsubscript{Single}} & 88.30  \small ± 0.10& 81.41  \small ± 1.49& \textbf{9.77  \small ± 0.28}& 8.41 \small  ± 0.47& \textbf{34.11  \small ± 0.73}& 76.84  \small ± 0.42\\
 \multicolumn{2}{l|}{OE\textsubscript{Ensemble}} & 89.61  \small ± 0.04& \textbf{84.53 \small  ± 0.72}& 9.93  \small ± 0.11& \textbf{8.35 \small  ± 0.20}& 34.34 \small  ± 0.39& \textbf{80.65 \small  ± 0.30}\\
 \multicolumn{2}{l|}{\textbf{SDDE\textsubscript{OOD}}} & \textbf{89.70  \small ± 0.22}& 85.47  \small ± 1.76& 10.13  \small ± 0.28& 8.70 \small  ± 0.31& 34.86 \small  ± 0.60& 80.30  \small ± 0.25\\ \hline
 & & \multicolumn{6}{c}{\bf ImageNet-200}\\ \hline
 \multicolumn{2}{l|}{OE\textsubscript{Single}} & 81.87  \small ± 0.16& 86.77  \small ± 0.06& 21.88  \small ± 0.46& 37.31 \small  ± 1.10& 54.67 \small  ± 1.11& 77.57  \small ± 0.28\\
 \multicolumn{2}{l|}{OE\textsubscript{Ensemble}} & 84.41  \small ± 0.15& 89.45  \small ± 0.26& 21.71  \small ± 0.07& \textbf{36.68  \small ± 0.14}& 54.37 \small  ± 0.12& \textbf{83.36 \small  ± 0.07}\\
 \multicolumn{2}{l|}{\textbf{SDDE\textsubscript{OOD}}} & \textbf{84.46  \small ± 0.07}& \textbf{89.57  \small ± 0.11}& \textbf{21.63 \small  ± 0.29}& 36.71 \small  ± 0.79& \textbf{54.17  \small ± 0.69}& \textbf{83.36  \small ± 0.08}\\ \bottomrule
\end{tabular}%
}
\label{tab:oe-new}
\end{table}

\subsection{Logit Aggregation Ablation}
\label{sec:logits_sdde_odd}

In Section \ref{sec:method}, we introduced logit-based prediction aggregation for the proposed SDDE approach. Although a single-model maximal logit score can be used for OOD detection \cite{hendrycks2019mls}, it is still unclear whether logit averaging is applicable for ensemble OOD detection. To determine this, we analyze the outputs of individual models in the ensemble. 

\begin{figure}[h!]
\centerline{\includegraphics[width=0.75\linewidth]{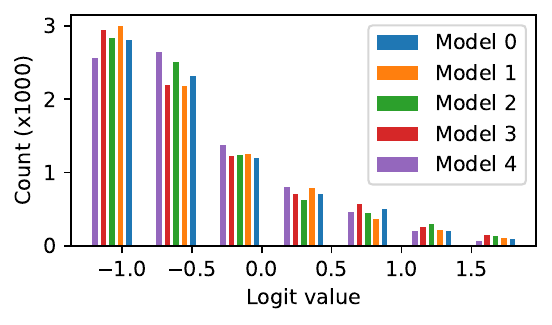}}
\caption{Logit distribution for individual
models in the SDDE ensemble for the first
class in CIFAR100.}
\label{fig:logits}
\end{figure}

As demonstrated in Figure \ref{fig:logits}, the logits exhibit similar scales and distributions. Although the figure only presents data from the CIFAR100 dataset and a single class, similar patterns are observed for other datasets and classes. Based on this observation, it appears that individual models contribute equally to the final OOD score, potentially enhancing the robustness of the OOD detection score estimation.

As our baselines use probability averaging aggregation, we ablate logit aggregation for SDDE. In this experiment, we apply the proposed MAL aggregation to all baseline methods. The results are presented in Table \ref{tab:aggr-abl}. Our findings show that MAL improves the quality of the baseline models. At the same time, SDDE achieves higher scores in 6 out of 8 comparisons. 

\begin{table}[h]
\caption{Near / Far OOD detection AUROC comparison for all methods with MAL aggregation.}
\label{tab:aggr-abl}
\centering
\resizebox{\linewidth}{!}{%
\begin{tabular}{@{}c|ccccc@{}}
\toprule
\textbf{Dataset} & \textbf{DE} & \textbf{NCL} & \textbf{ADP} & \textbf{DICE} & \textbf{SDDE} \\ \hline
MNIST & 93.45 / 99.33 & 93.69 / 99.35 & 93.90 / 99.28 & 93.88 / 99.24 & \textbf{96.24 / 99.70} \\
CIFAR10 & 91.66 / 94.36 & 91.67 / 94.24 & 91.08 / 94.02 & 90.44 / 93.64 & \textbf{91.90 / 94.55} \\
CIFAR100 & 83.36 / 82.03 & 83.63 / 81.89 & 83.04 / 81.55 & 83.62 / \textbf{83.40} & \textbf{83.69} / 82.39 \\
ImageNet & 75.21 / 85.80 & 74.89 / 85.88 & {\bf75.82} / 84.03 & N/A & 75.26 / \bf 85.98 \\
\bottomrule
\end{tabular}%
}
\end{table}

\section{Related Work}

\textbf{Confidence Estimation.}
DNNs can be prone to overfitting, which limits their ability to generalize and predict confidence \cite{guo2017calibration}. Multiple works attempted to address this issue. A simple approach to confidence estimation is to take the probability of a predicted class on the output of the softmax layer \cite{hendrycks2016msp}. Several authors proposed improvements to this method for either better confidence estimation or higher OOD detection accuracy.

Some works studied activation statistics between layers to detect anomalous behavior \cite{sun2021react}. Another approach is to use Bayesian training, which can lead to improved confidence prediction at the cost of accuracy \cite{goan2020bayesian}. Other works attempted to use insights from classical machine learning, such as KNNs \cite{yang2022openood}
and ensembles \cite{lakshminarayanan2017deepensemble}. It was shown that ensembles reduce overfitting and produce confidence estimates, which can outperform most existing methods \cite{yang2022openood}. In this work, we further improve deep ensembles by applying a new training algorithm.

\textbf{Ensembles Diversification.}
According to previous works, ensemble methods produce SOTA results on popular classification \cite{lakshminarayanan2017deepensemble, rame2021dice} and OOD detection tasks \cite{yang2022openood}. It was shown that the quality of the ensemble largely depends on the diversity of underlying models \cite{stickland2020diverseaug}. Some works improve diversity by implementing special loss functions on predicted probabilities. The idea is to make different predicted distributions for different models. The NCL loss \cite{shui2018ncl} reduces correlations between output probabilities, but can negatively influence the prediction of the correct class. The ADP \cite{pang2019adp} solves this problem by diversifying only the probabilities of alternative classes. On the other hand, the DICE \cite{rame2021dice} loss reduces dependency between bottleneck features of multiple models.
While previous works reduce the correlations of either model predictions or bottleneck features, we directly diversify the input features used by the models within an ensemble, which further improves OOD detection and calibration.

\textbf{Out-of-Distribution Detection.}
Risk-controlled recognition poses a problem for detecting out-of-distribution (OOD) data, i.e., data with distribution different from the training set, or data with unknown classes \cite{yang2022openood}. Multiple OOD detection methods were proposed for deep models \cite{sun2021react,hendrycks2019mls}. Despite the progress made on single models, Deep Ensembles \cite{lakshminarayanan2017deepensemble} use a traditional Maximum Softmax Probability (MSP) \cite{hendrycks2016msp} approach to OOD detection. In this work, we introduce a novel ensembling method that offers enhanced OOD detection capabilities. Furthermore, we extend the Maximum Logit Score (MLS) \cite{hendrycks2019mls} for ensemble application, underscoring its advantages over MSP.

\section{Limitations}

Incorporating OOD samples during ensemble training can be beneficial, but it raises concerns about sample quality, diversity, and source, and whether results will generalize to unseen OOD categories. Achieving SOTA results on the OpenOOD benchmark is notable, but benchmarks evolve, and the method's real-world applicability and robustness across varied scenarios and datasets need further testing. These limitations qualify, but do not invalidate, our conclusions, offering starting points for future research.

\section{Conclusion}

In this work, we proposed SDDE, a novel ensembling method for classification and OOD detection. SDDE forces the models within the ensemble to use different input features for prediction, which increases ensemble diversity. According to our experiments, SDDE performs better than several popular ensembles on the CIFAR10, CIFAR100, and ImageNet-1K datasets. At the same time, SDDE outperforms other methods in OOD detection on the OpenOOD benchmark. Improved confidence estimation and OOD detection make SDDE a valuable tool for risk-controlled recognition. We further generalized SDDE for training with OOD data by proposing SDDE\textsubscript{OOD} approach. SDDE\textsubscript{OOD} achieves SOTA results on the OpenOOD benchmark.

\bibliographystyle{IEEEbib}
\bibliography{strings,refs}

\begin{thebibliography}{10}

\bibitem{liu2020detection}
Li~Liu et~al.,
\newblock ``Deep learning for generic object detection: A survey,''
\newblock {\em International journal of computer vision}, vol. 128, pp. 261--318, 2020.

\bibitem{he2015relu}
Kaiming He et~al.,
\newblock ``Delving deep into rectifiers: Surpassing human-level performance on imagenet classification,''
\newblock in {\em Proceedings of the IEEE international conference on computer vision}, 2015, pp. 1026--1034.

\bibitem{deng2019arcface}
Jiankang Deng et~al.,
\newblock ``Arcface: Additive angular margin loss for deep face recognition,''
\newblock in {\em Proceedings of the IEEE/CVF conference on computer vision and pattern recognition}, 2019, pp. 4690--4699.

\bibitem{malinin2022shifts}
Andrey Malinin et~al.,
\newblock ``Shifts 2.0: Extending the dataset of real distributional shifts,'' 2022.

\bibitem{koh2020wilds}
Pang~Wei Koh et~al.,
\newblock ``{WILDS}: A benchmark of in-the-wild distribution shifts,''
\newblock {\em arXiv}, 2020.

\bibitem{yang2022openood}
Jingyang Zhang et~al.,
\newblock ``Openood v1.5: Enhanced benchmark for out-of-distribution detection,''
\newblock {\em arXiv preprint arXiv:2306.09301}, 2023.

\bibitem{guo2017calibration}
Chuan Guo et~al.,
\newblock ``On calibration of modern neural networks,''
\newblock in {\em International conference on machine learning}. PMLR, 2017, pp. 1321--1330.

\bibitem{goan2020bayesian}
Ethan Goan et~al.,
\newblock ``Bayesian neural networks: An introduction and survey,''
\newblock {\em Case Studies in Applied Bayesian Data Science: CIRM Jean-Morlet Chair, Fall 2018}, pp. 45--87, 2020.

\bibitem{lakshminarayanan2017deepensemble}
Balaji Lakshminarayanan et~al.,
\newblock ``Simple and scalable predictive uncertainty estimation using deep ensembles,''
\newblock {\em Advances in neural information processing systems}, vol. 30, 2017.

\bibitem{shui2018ncl}
Changjian Shui et~al.,
\newblock ``Diversity regularization in deep ensembles,''
\newblock {\em arXiv preprint arXiv:1802.07881}, 2018.

\bibitem{pang2019adp}
Tianyu Pang et~al.,
\newblock ``Improving adversarial robustness via promoting ensemble diversity,''
\newblock in {\em International Conference on Machine Learning}. PMLR, 2019, pp. 4970--4979.

\bibitem{rame2021dice}
Alexandre Ram{\'e} et~al.,
\newblock ``Dice: Diversity in deep ensembles via conditional redundancy adversarial estimation,''
\newblock in {\em ICLR 2021-9th International Conference on Learning Representations}, 2021.

\bibitem{hendrycks2018oe}
Dan Hendrycks et~al.,
\newblock ``Deep anomaly detection with outlier exposure,''
\newblock {\em arXiv preprint arXiv:1812.04606}, 2018.

\bibitem{simonyan2013saliencegrad}
Karen Simonyan et~al.,
\newblock ``Deep inside convolutional networks: Visualising image classification models and saliency maps,''
\newblock {\em arXiv preprint arXiv:1312.6034}, 2013.

\bibitem{zhou2016cam}
Bolei Zhou et~al.,
\newblock ``Learning deep features for discriminative localization,''
\newblock in {\em Proceedings of the IEEE conference on computer vision and pattern recognition}, 2016, pp. 2921--2929.

\bibitem{selvaraju2017gradcam}
Ramprasaath~R Selvaraju et~al.,
\newblock ``Grad-cam: Visual explanations from deep networks via gradient-based localization,''
\newblock in {\em Proceedings of the IEEE international conference on computer vision}, 2017, pp. 618--626.

\bibitem{hendrycks2016msp}
Dan Hendrycks et~al.,
\newblock ``A baseline for detecting misclassified and out-of-distribution examples in neural networks,''
\newblock {\em arXiv preprint arXiv:1610.02136}, 2016.

\bibitem{abe2022necessary}
Taiga Abe et~al.,
\newblock ``Deep ensembles work, but are they necessary?,''
\newblock {\em Advances in Neural Information Processing Systems}, vol. 35, pp. 33646--33660, 2022.

\bibitem{hendrycks2019mls}
Dan Hendrycks et~al.,
\newblock ``Scaling out-of-distribution detection for real-world settings,''
\newblock {\em arXiv preprint arXiv:1911.11132}, 2019.

\bibitem{he2016resnet}
Kaiming He et~al.,
\newblock ``Deep residual learning for image recognition,''
\newblock in {\em Proceedings of the IEEE conference on computer vision and pattern recognition}, 2016, pp. 770--778.

\bibitem{krizhevsky2009cifar}
Alex Krizhevsky et~al.,
\newblock ``Learning multiple layers of features from tiny images,''
\newblock Tech. {R}ep.~0, University of Toronto, Toronto, Ontario, 2009.

\bibitem{lecun1998mnist}
Yann LeCun et~al.,
\newblock ``Gradient-based learning applied to document recognition,''
\newblock {\em Proceedings of the IEEE}, vol. 86, no. 11, pp. 2278--2324, 1998.

\bibitem{loshchilov2016cosineann}
Ilya Loshchilov et~al.,
\newblock ``Sgdr: Stochastic gradient descent with warm restarts,''
\newblock {\em arXiv preprint arXiv:1608.03983}, 2016.

\bibitem{aksela2003comparison}
Matti Aksela,
\newblock ``Comparison of classifier selection methods for improving committee performance,''
\newblock in {\em International Workshop on Multiple Classifier Systems}. Springer, 2003, pp. 84--93.

\bibitem{brier1950verification}
Glenn~W Brier et~al.,
\newblock ``Verification of forecasts expressed in terms of probability,''
\newblock {\em Monthly weather review}, vol. 78, no. 1, pp. 1--3, 1950.

\bibitem{hendrycks2019robustness}
Dan Hendrycks et~al.,
\newblock ``Benchmarking neural network robustness to common corruptions and perturbations,''
\newblock {\em Proceedings of the International Conference on Learning Representations}, 2019.

\bibitem{sun2021react}
Yiyou Sun et~al.,
\newblock ``React: Out-of-distribution detection with rectified activations,''
\newblock {\em Advances in Neural Information Processing Systems}, vol. 34, pp. 144--157, 2021.

\bibitem{stickland2020diverseaug}
Asa~Cooper Stickland et~al.,
\newblock ``Diverse ensembles improve calibration,''
\newblock {\em arXiv preprint arXiv:2007.04206}, 2020.

\end{thebibliography}

\end{document}